\documentclass[11pt,a4paper]{article}
\usepackage[hyperref]{acl2021}
\usepackage[utf8]{inputenc}
\usepackage[T2A,T1]{fontenc}
\usepackage[russian,english]{babel}
\usepackage{longtable}

\usepackage{times}
\usepackage{latexsym}
\usepackage{graphicx}
\usepackage{booktabs}

\usepackage{enumitem}

\usepackage{microtype}

\aclfinalcopy 

\newcommand{\R}{\textit{RuShiftEval}}

\newcommand{\RU}[2]{\foreignlanguage{russian}{#1} (`#2')}

\newcommand{\nevlezlo}[1]{}

\title{Three-part diachronic semantic change dataset for Russian}
\author{Andrey Kutuzov \\
  University of Oslo \\
  Norway \\
  \texttt{andreku@ifi.uio.no} \\\And
  Lidia Pivovarova \\
  University of Helsinki \\
  Finland \\
  \texttt{lidia.pivovarova@helsinki.fi} \\}

\date{}

\begin{document}
\maketitle
\begin{abstract}
We present a manually annotated lexical semantic change dataset for Russian: RuShiftEval. Its novelty is ensured by a single set of target words annotated for their diachronic semantic shifts across three time periods, while the previous work either used only two time periods, or different sets of target words. The paper describes the composition and annotation procedure for the dataset. In addition, it is shown how the ternary nature of RuShiftEval allows to trace specific diachronic trajectories: `changed at a particular time period and stable afterwards' or `was changing throughout all time periods'. Based on the analysis of the submissions to the recent shared task on semantic change detection for Russian, we argue that correctly identifying such trajectories can be an interesting sub-task itself.
\end{abstract}

\section{Introduction}

This paper describes \R{}: a new dataset of diachronic semantic changes for Russian words. Its novelty in comparison with prior work is its multi-period nature. Until now, semantic change detection datasets focused on shifts occurring between \textbf{two} time periods\nevlezlo{(for example, `1810-1860' versus `1960-2010' in the English subset of the SemEval 2020 Task 1 dataset \cite{schlechtweg-etal-2020-semeval})}. On the other hand, \R{} provides human-annotated degrees of semantic change for a set of Russian nouns over \textbf{three} time periods: pre-Soviet (1700-1916), Soviet (1918-1990) and post-Soviet (1992-2016).  Notably, it also contains `skipping' comparisons of pre-Soviet meanings versus post-Soviet meanings. Together, this forms three subsets: \R{}-1 (pre-Soviet VS Soviet), \R{}-2 (Soviet VS post-Soviet) and \R{}-3 (pre-Soviet VS post-Soviet).

The three periods naturally stem from the Russian history: they were radically different in terms of life realities and writing and practices, which is reflected in the language.  As an example, the word \foreignlanguage{russian}{дядька} lost its `tutor of a kid in a rich family' sense in the Soviet times, with only the generic `adult man' sense remaining.
Certainly, language development never stops and Russian also gradually evolved within those periods as well, not only on their boundaries. However, in order to create a usable semantic change dataset, one has to draw the boundaries somewhere, and it is difficult to come up with more fitting `changing points' for Russian.
\nevlezlo{Note, that the pre-Soviet data consist mostly of the 19th century texts. }

\R{} can be used for testing the ability of semantic change detection systems to trace long-term multi-point dynamics of diachronic semantic shifts, rather than singular change values measured by comparing two time periods. As such, \R{} was successfully employed in a recent shared task on semantic change detection for Russian \cite{rushifteval2021}.

\section{Related work}

Automatic detection of word meaning change is a fast growing research area~\cite{kutuzov-etal-2018-diachronic,tahmasebi2018survey}. Evaluation of this task is especially challenging; \textit{inter alia}, it requires gold standard annotation covering multiple word usages.
\nevlezlo{
There are some attempts to evaluate systems without explicit data annotation, for example, using known distinctions between nouns and verbs~\cite{hamilton-etal-2016-cultural}. Another idea is to use synthetic datasets, where distribution of word meaning could be controlled by either merging two words together~\cite{tsakalidis-liakata-2020-sequential,shoemark-etal-2019-room,rosenfeld-erk-2018-deep} or using a sense-annotated corpus~\cite{schlechtweg2020simulating}. Nevertheless, semantic change regularities are yet to be studied and it is very important to annotate actual meaning change cases. }

The common practice is to annotate pairs of sentences as using a target word in either the same or different senses. It was introduced for the word sense disambiguation task in~\cite{erk-etal-2013-measuring}, while \cite{schlechtweg-etal-2018-diachronic} proposed methods to aggregate pairwise annotations for semantic change modeling; one of them, the COMPARE metrics, is used in \R{}. 

A similar approach was used for the SemEval'20 shared task on semantic change detection~\cite{schlechtweg-etal-2020-semeval}: annotators labeled pairs of sentences, where some pairs belonged to the same periods and some to different ones. This annotation resulted in a diachronic word usage graph, which was then clustered to obtain separate word senses and their distributions between time periods~\cite{schlechtweg2021dwug}.  

The pairwise sentence annotation has been used in creating another semantic change dataset for Russian, \textit{RuSemShift}~\cite{rodina-kutuzov-2020-rusemshift}. We use the same annotation procedure and rely on the same corpus, i.e. Russian National Corpus (RNC) split into pre-Soviet, Soviet and post-Soviet sub-corpora. However, \textit{RuSemShift} features two sets of words: one for the changes between the pre-Soviet and Soviet periods, and another for the Soviet and post-Soviet periods. The new \R{} dataset, which we present in this paper, uses a \textit{joint word set} allowing for tracing each word across three time periods. In addition, we directly annotate semantic change between the pre-Soviet and post-Soviet periods, skipping the Soviet one.

\section{Dataset Construction}

\subsection{Word List Creation}

In building the dataset, we relied on the graded view on word meaning change~\cite{schlechtweg2021dwug}: for each word in the dataset, we measure a \textit{degree of change} between pairs of periods, rather than making a binary decision on whether its sense inventory changed over time. The measure relies on pairwise sentence annotations, where each pair of sentences is processed by at least three annotators.

Compiling the target-word set, we needed to ensure two main conditions: (i) the dataset contains many `interesting' words, i.e. words that changed their meaning between either pair of periods; (ii) not all words in the dataset actually changed their meaning. We followed the same procedure as in~\cite{kutuzov2018two,rodina-kutuzov-2020-rusemshift,schlechtweg-etal-2020-semeval}: first, select changing words, and then augment them with \textit{fillers}, i.e. random words following similar frequency distribution across three time periods.

Technically, it was possible to populate the target word set automatically, using any pre-trained language model (LM) for Russian and some measure of distance between word representations in different corpora. However, we wanted our target words choice to be motivated linguistically rather than influenced by any LM architecture. Therefore, to find changing words, we first consulted several dictionaries of outdated or, on the contrary, the most recent Russian words, such as~\cite{novikov2016,basko2011,tolk1998}. Unfortunately, dictionaries provided less examples than we needed: they often contain archaisms, neologisms, multi-word expressions, and words which are infrequent in the corpus or not used in the meanings specified in the dictionaries. 

However, we discovered that some changing words could be found in papers on specific linguistics problems. For example, the word \RU{облако}{cloud} was found in a paper on the Internet language~\cite{baldanova2016}; \RU{стол}{table/diet}---in an article discussing the language of one story by Pushkin~\cite{elmi2016}. Finally, to find some of the target words, we used our intuition as educated native speakers. Out of 50 words, 13 were found in dictionaries, 10 invented by ourselves and the rest 27 found in articles on more specific topics.

Regardless the initial word origin, we manually checked  that all words occur at least 50 times in each of the three sub-corpora and that the distinctive sense is used several times.

Fillers (selected for each target word) are sampled so that they belong to the same part of speech---nouns in our case---and their frequency percentile is the same as the target word frequency percentile in all three periods. The aim here is to ensure that frequency cannot be used to distinguish the target words from fillers.\footnote{Indeed, there is no significant correlation between frequency differences and the aggregated relatedness scores from our gold annotation.}
For \R{}, we sampled two filler words for each target word.

The final dataset consists of 111 Russian nouns, where 12 words form a development set and 99 words serve as a test set. Since the annotation procedure is the same as for \textit{RuSemShift} \cite{rodina-kutuzov-2020-rusemshift}, one can use one of these resources as a training set and then evaluate on another\nevlezlo{ (\textit{RuSemShift} features two subsets, with 48 and 51 target words)}.

\subsection{Annotation} \label{subsec:ann}
Annotators' guidelines were identical to those in \textit{RuSemShift} \cite{rodina-kutuzov-2020-rusemshift}.
To generate annotation tasks,
we sampled 30 sentences from each sub-corpus and created sentence pairs. We ran this sampling independently for all three period pairs. The sentences were accompanied by one preceding and one following sentence, to ease the annotators' work in case of doubt. The task was formulated as labeling on a 1-4 scale, where $1$ means the senses of the target word in two sentences are unrelated, $2$ stands for `distantly related', $3$ stands for `closely related', and $4$ stands for `senses are identical' \cite{hatty-etal-2019-surel}.  Annotators were also allowed to use the $0$ (`cannot decide') judgments. They were excluded from the final datasets, but their number was negligible anyway: about 100 out of total 30 000.

The annotation was carried out on the Yandex.Toloka crowd-sourcing platform.\footnote{\url{https://toloka.yandex.ru/}}
We employed native speakers of Russian, older than 30, with a university degree.  To ensure the annotation quality, the authors themselves annotated about 20 control examples for each pair of periods. We chose the most obvious cases of $1$ and $4$ for this; annotators who answered incorrectly (not with the exactly matching grade), were banned from the task for 24 hours. The inter-rater agreement statistics and the number of judgments in each \R{} subset are shown in Table~\ref{tab:stats}. The agreement is on par with other semantic change annotation efforts: \cite{schlechtweg-etal-2020-semeval} report Spearman correlations ranging from $0.58$ to $0.69$, \cite{rodina-kutuzov-2020-rusemshift} report Krippendorff’s $\alpha$ ranging from $0.51$ to $0.53$.\footnote{Note it does not make much sense to report correlations for individual annotators (`data columns'), since in our crowd-working setup, the columns are not associated with particular persons.} Each subset was annotated by about 100 human raters, more or less uniformly `spread' across annotation instances, with the only constraint being that each instance must be annotated by three different persons. 

\begin{table}
    \centering
    \begin{tabular}{l|cccc}
    \textbf{Time bins} & $\alpha$ & $\rho$ &  \textbf{JUD} & \textbf{0-JUD} \\ 
    \midrule
    \multicolumn{5}{c}{Test set (99 words)} \\
    \midrule
    RuShiftEval-1 & 0.506 & 0.521 & 8 863 & 42 \\
    RuShiftEval-2 & 0.549 & 0.559 & 8 879 & 25 \\
    RuShiftEval-3 & 0.544 & 0.556 & 8 876 & 31 \\
    \midrule
    \multicolumn{5}{c}{Development set (12 words)} \\
    \midrule
    RuShiftEval-1 & 0.592  & 0.613  & 1 013 & 7  \\
    RuShiftEval-2 & 0.609 & 0.627  & 1 014 & 3  \\
    RuShiftEval-3 & 0.597  & 0.632  & 1 015  & 2\\
    \bottomrule
    \end{tabular}
    \caption{\R{} statistics. $\alpha$ and $\rho$ are inter-rater agreement scores as calculated by Krippendorff's $\alpha$ (ordinal scale) and mean pairwise Spearman $\rho$. JUD is total number of judgments and 0-JUD is the number of 0-judgments (`cannot decide').}
    \label{tab:stats}
\end{table}

Finally, the degrees of semantic change for each word between a pair of periods were calculated using the COMPARE metrics~\cite{schlechtweg-etal-2018-diachronic}, which is the average of pairwise relatedness scores. Interestingly, some words initially sampled as fillers---e.g. \RU{ядро}{cannonball or core/nucleus}---ended up among most changed according to the annotation.  Also some words from the initial set were annotated as relatively stable. This happened because the distinctive sense was rare or because annotators' opinion diverged from linguistic knowledge in the dictionaries. For example, for the word \RU{бригада}{brigade/gang/team} dictionaries list two distinct senses---a military and a civil one. However, in most cases the annotators considered these senses identical or closely related. 

The dataset is publicly available, including the raw scores assigned by annotators.\footnote{\url{https://github.com/akutuzov/rushifteval_public}} 

\begin{table*}[t!]
    \centering
    \begin{tabular}{lp{320px}|cc}
    \toprule
        \textbf{Type}  & \textbf{Examples} & \textbf{Baseline} & \textbf{Top}  \\
        \midrule
    1 & \foreignlanguage{russian}{закладка} (`foundation/bookmark/hidden artifact'), \foreignlanguage{russian}{линейка} (`carriage/ruler/series of goods'), \foreignlanguage{russian}{центр} (`center') & 0.5 & 1.0 \\
      \midrule
    2 & \foreignlanguage{russian}{дядька} (`tutor/adult man'), \foreignlanguage{russian}{живот} (`life/belly/stomach'), \foreignlanguage{russian}{лох} (`salmon/silver-berry/easy victim, stupid person'), \foreignlanguage{russian}{роспись} (`list/painting'), \foreignlanguage{russian}{ядро} (`cannonball/core/nucleus') & 1.0  & 1.0 \\
      \midrule
    3 & \foreignlanguage{russian}{полоса} (`stripe/ribbon/lane/runway'), \foreignlanguage{russian}{связка} (`ligament/vocal cords/mutual connection'), \foreignlanguage{russian}{спутник} (`fellow traveler/satellite/sputnik'), \foreignlanguage{russian}{ссылка} (`exile/link'), \foreignlanguage{russian}{тачка} (`wheelbarrow/car'), \foreignlanguage{russian}{формат} (`format') & 0.4 & 0.8-1.0 \\
    \bottomrule
    \end{tabular}
    \caption{Semantic change trajectory types in \R{} and the percentage of words with correctly captured type for the baseline and the 4 best shared task submissions (see \ref{subsec:eval}).}
    \label{tab:trajectories}
\end{table*}

\section{Diachronic trajectory types}

\R{}  allows tracing multi-hop dynamics of semantic change.\nevlezlo{ It is possible to imagine examples like a word $X$ keeping its meaning more or less intact during the pre-Soviet and post-Soviet periods, but then changing it in the post-Soviet period. In fact,} A similar analysis of diachronic word embedding series or `trajectories' was conducted in \cite{kulkarni2015statistically} and \cite{hamilton-etal-2016-diachronic}, but the former focused on change point detection, and the latter on finding general laws of semantic change. With manually annotated \R{} dataset we were able to move further and identify at least three different types of changing trajectories: 1) changes in every period pair; 2) change in the Soviet period as compared to the pre-Soviet period; 3) change in the post-Soviet period as compared to the Soviet period. 

Since approximately a half of the words in the dataset did not change their meaning they exhibit a fourth, trivial type of trajectory, where all three distances are small. In principle there could be a fifth type of trajectory, where difference between pre-Soviet and post-Soviet periods is substantially smaller than between other period pairs, which would mean that a word was used in a new sense during the Soviet time but then came back to its original meaning. However, we did not find any words following this trajectory type and not sure whether this behavior is theoretically plausible.

Table~\ref{tab:trajectories} shows examples of nouns belonging to three non-stable trajectory types. Below we explain the semantic change processes for them.

\paragraph{1.} The word \foreignlanguage{russian}{закладка} belongs to the type 1. Its dominant sense in the pre-Soviet period was `foundation' (as in `\textit{The foundation of the new church building took place yesterday}'). In the Soviet times, the `bookmark' sense emerged (it was already present before, but very rare). Then, the post-Soviet time period saw the emergence of two new senses, both through widening processes: `tab' (in graphical user interfaces) and `booby-trapping' or `something hidden' (often about illegal drugs cached by a distributor). Thus, low relatedness scores are observed across all possible pairs: the word is used differently in each time period.

\paragraph{2.} The word \foreignlanguage{russian}{ядро} can mean either `cannonball' or `core/kernel/nucleus'. It belongs to the type 2. In the Soviet period, the first sense almost disappeared (because artillery stopped using cannonballs in the 20\textsuperscript{th} century), while the latter sense became more frequent. After this reduction, the meaning was stable, with no changes in the post-Soviet period\nevlezlo{ (as a result, the relatedness score between the post-Soviet and the pre-Soviet periods is not surprisingly low)}. 

\paragraph{3.} The word \foreignlanguage{russian}{тачка} (`wheelbarrow') belongs to the type 3. It was stable until the end of the Soviet period, but in the post-Soviet times, \foreignlanguage{russian}{тачка} acquired a new colloquial sense of `car', quite common even in written texts. This lead to divergence from both Soviet and pre-Soviet periods. 

\begin{figure}
    \centering
    \includegraphics[width=0.3\linewidth]{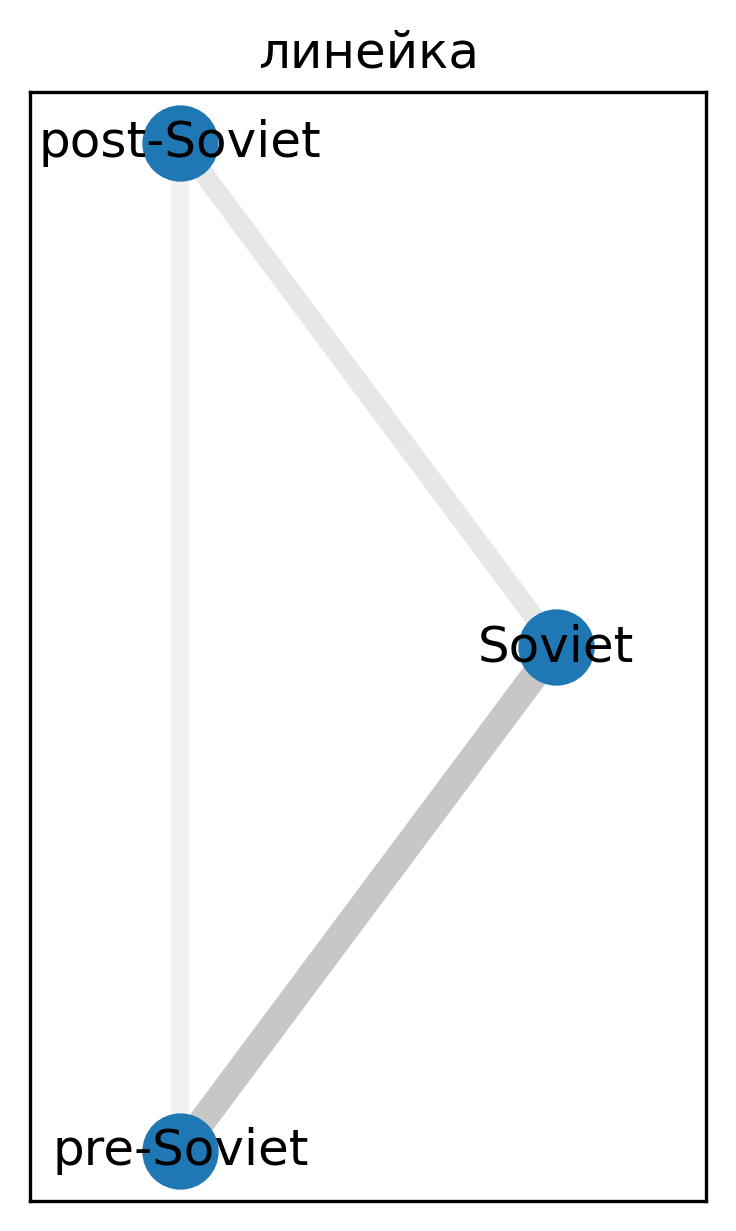}
    \includegraphics[width=0.3\linewidth]{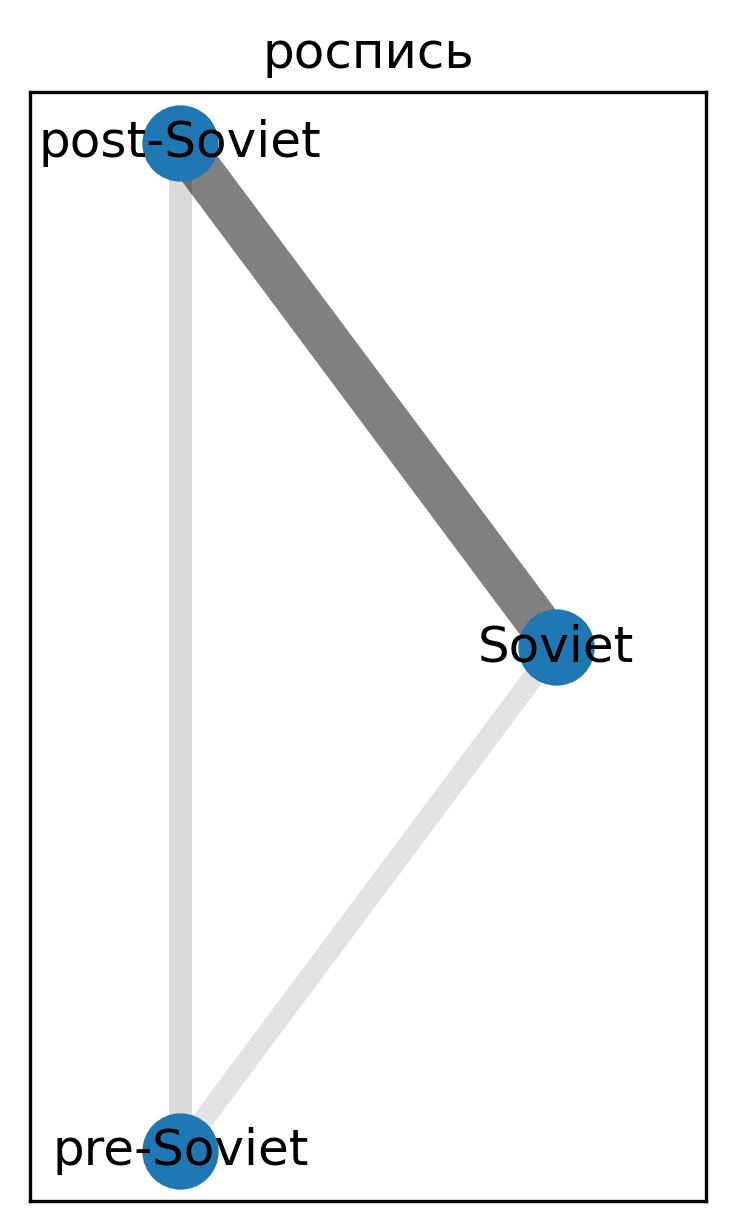}
    \includegraphics[width=0.3\linewidth]{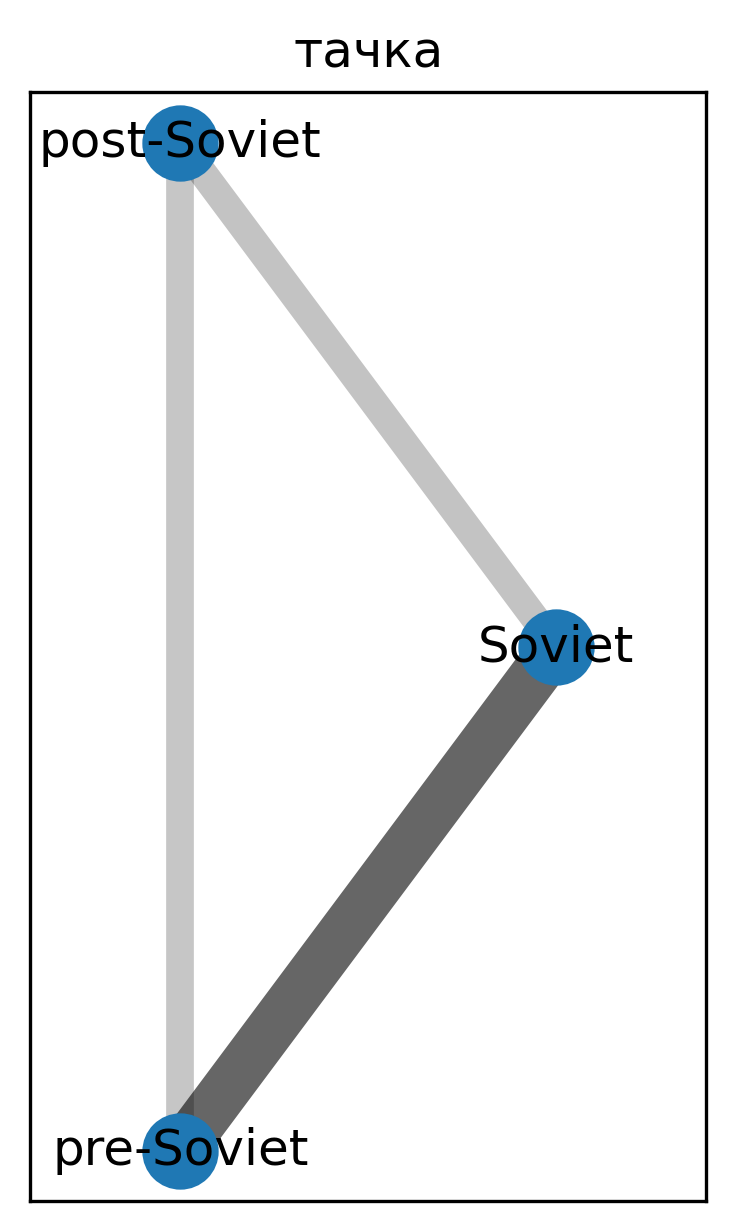}
    \caption{Time relatedness graphs for words belonging to different semantic trajectory types (from left to right): \foreignlanguage{russian}{линейка} (`carriage/ruler/series of goods') (1), \foreignlanguage{russian}{роспись} (`list/painting') (2), \foreignlanguage{russian}{тачка} (`wheelbarrow/car') (3).}
    \label{fig:tachka}
\end{figure}

Semantic trajectory types could be visualized as time relatedness graphs; see Figure~\ref{fig:tachka}. Nodes of the graph are time periods, and edge widths represent the COMPARE score (see \ref{subsec:ann}) for each pair of periods.\footnote{Note that in most cases it is impossible to use nodes relative positions on the plot to reflect relatedness scores: one can't change the length of an edge in a triangle without also changing the length of at least one other edge.} Thus, thicker edges denote stable meaning, while thinner and more transparent edges show a change. Each trajectory type has its own characteristic pattern of edge widths. For example, in the graph for \foreignlanguage{russian}{тачка} (the rightmost plot), the edges connecting the post-Soviet node to two other nodes are much thinner than the edge between the pre-Soviet and post-Soviet nodes. This signals a change in the post-Soviet times (trajectory type 3).

Note that the annotation process and the definition of the COMPARE score itself do not guarantee perfect capturing of semantic changes. One example---made clear by the multi-period nature of \R{} design---is the word \foreignlanguage{russian}{радикал} (`radical'). Its relatedness scores are low across all time period pairs, suggesting that it experienced sequential changes similar to \foreignlanguage{russian}{закладка}. However, in fact, throughout all the times covered by \R{}, this word had the same two persistent senses: political and chemical. Since their probabilities were almost equal, many randomly sampled sentence pairs contained the word \foreignlanguage{russian}{радикал} in two different senses, which led to low COMPARE scores. In this case, it stems from strong and persistent ambiguity of the word, not from diachronic semantic change. This limitation of the COMPARE metrics was already described in \cite{schlechtweg-etal-2018-diachronic}\nevlezlo{, so it is expected, but still should be taken into account}.

Another potential flaw is sampling variability. For annotation, we sampled 30 sentences with a target word from each time period for each comparison. Since our relatedness graph has three edges, each word is represented with two samples. As it turned out, in some cases different samples can yield quite different picture of sense distributions. 

Let us manually analyze the word \foreignlanguage{russian}{полость} (`cavity/hide to cover one's legs in an open cart'). Since horse-driven carts disappeared just a few years after the beginning of the Soviet period, one might expect the second sense to be lost in Soviet times and never to appear again. However, the relatedness between the Soviet and post-Soviet time periods ($1.9$) is even lower than between the pre-Soviet and Soviet periods ($2.2$), as if the word experienced another semantic shift. In fact, it is a random sampling artifact. In the 30 sentences from the Soviet period sampled for the `pre-Soviet:Soviet' pair, only 4 used \foreignlanguage{russian}{полость} in this archaic sense. But in the 30 sentences \textit{from the same period} sampled for the `Soviet:post-Soviet' pair, this number grew to 10, $2.5$ times more (mostly in fiction texts, where the plot is set in the pre-Soviet times). As a result, the Soviet usage pattern looks like it is different from the post-Soviet one, although in fact no shift has happened (as evident both from linguistic intuition of Russian speakers and from the Fisher exact test which in this case yields $p=0.13$). The frequency of \foreignlanguage{russian}{полость} in the Soviet sub-corpus is about 600, so both samples together cover only 10\% of the full concordance. Without manually annotating all six hundred occurrences, it is difficult to tell which sample is more representative of the real word usage in the Soviet times. \nevlezlo{The take-home message here is that when using a similar annotation workflow, one must try} It would be better to increase the sample size as much as possible: 30 is arguably already on the border.\nevlezlo{ But of course, material resources for annotation enforce their limits.}

\subsection{Trajectory detection task?} \label{subsec:eval}
The \R{} dataset was used to evaluate the systems participating in a shared task on lexical semantic change detection for Russian \cite{rushifteval2021}. How good these submissions are in capturing the trajectory types described in the previous section? In this subsection, we describe a toy experiment to address this question.

For simplicity, we will use only 11 example words from Table~\ref{tab:trajectories} which appear in the \R{} evaluation set (this excludes \foreignlanguage{russian}{закладка}, \foreignlanguage{russian}{лох} and \foreignlanguage{russian}{спутник}, since they appear in the development set only). Then a set of criteria is established for the system predictions, corresponding to each of the three trajectory types.
We consider a system successful in capturing a word with the \textbf{trajectory 2} if the predicted relatedness score is higher for the `Soviet:post-Soviet' pair than for other two pairs. For the words with the \textbf{trajectory 3}, the relatedness score for the `pre-Soviet:Soviet' pair must be the highest among all pairs. For the words with the \textbf{trajectory 1}, the percentile ranks of the relatedness scores for all three sub-sets must be below $50$ (admittedly, this is an \textit{ad hoc} criterion, but it is used here just to give an example of how the task can be set up). Thus, at least for the trajectory types 2 and 3, this resembles a simple ranking task: not across target words within one period pair, but for one target word across three period pairs. At the same time, the trajectory type 1 (changes in every period) does not quite fit into this frame.

We compared the baseline system (which used static diachronic word embeddings and the local neighbors method from \cite{hamilton-etal-2016-cultural}) and four best systems (employing contextualized language models: ELMo, BERT or XLM-R). The results are presented in Table~\ref{tab:trajectories}. All of the best submissions captured the \textbf{trajectory 1} for all two target words, but the baseline method failed for \foreignlanguage{russian}{центр} (its percentile rank in \R{}-1 is more than $60$). For the \textbf{trajectory 3}, the top systems are considerably better than the baseline method. For example, according to the baseline method, \foreignlanguage{russian}{полоса} experienced its strongest change in the Soviet times, while in fact it was in the post-Soviet period. Only for the \textbf{trajectory 2}, the baseline is on par with the winners of the shared task.

This analysis is rather preliminary, but it shows that the systems performance in correctly detecting diachronic trajectories does to some extent correlate with their performance in the `traditional' semantic change ranking (with binary datasets, like in the SemEval 2020 Shared Task 1). We believe that this can be an interesting sub-task within the larger field of semantic change detection, once more datasets like \R{} are available and more formal definitions of `capturing the trajectory successfully' are developed. 

\section{Conclusion}

We presented \R{}, a novel dataset of diachronic semantic changes in Russian nouns across three time periods, using the same set of target words for all comparisons. We also conducted a preliminary analysis of how \R{} can be used in tracing diachronic semantic trajectories, and how current change detection systems for Russian deal with this potentially interesting task. 

\section*{Acknowledgments}
The annotation effort for \R{} was supported by the Russian Science Foundation grant 20-18-00206.  
This work has been partially supported by the European Union Horizon 2020 research and innovation programme under grants 770299 (NewsEye) and 825153 (EMBEDDIA).

\bibliographystyle{acl_natbib}
\bibliography{anthology,rushifteval}

\newpage
\appendix
\onecolumn
\newpage
\section{Transliterations of Russian words mentioned in the article}
\label{sec:dataset}

\begin{table*}[h]
    \centering
    \begin{tabular}{|l|l|l|}
    \toprule
\textsc{\textbf{word}} & \textsc{\textbf{transliteration}} & \textsc{\textbf{translation}} \\\midrule
\foreignlanguage{russian}{бригада} & brigada & brigade/gang/team \\\midrule
\foreignlanguage{russian}{дядька} & djadka & uncle/man/(male) tutor \\\midrule
\foreignlanguage{russian}{живот} & život & stomach/belly/life \\\midrule
\foreignlanguage{russian}{закладка} & zakladka & foundation/bookmark/hidden artifact \\\midrule
\foreignlanguage{russian}{линейка} & lineika & carriage/ruler/series of goods \\\midrule
\foreignlanguage{russian}{лох} & loh & salmon/silver-berry/easy victim \\\midrule
\foreignlanguage{russian}{облако} & oblako & cloud \\\midrule
\foreignlanguage{russian}{полоса} & polosa & tripe/ribbon/lane/runway \\\midrule
\foreignlanguage{russian}{полость} & polost & cavity/foot hide \\\midrule
\foreignlanguage{russian}{радикал} & radikal & radical \\\midrule
\foreignlanguage{russian}{роспись} & rospis & mural/signature/list \\\midrule
\foreignlanguage{russian}{связка} & svjazka & ligament/vocal cords/mutual connection \\\midrule
\foreignlanguage{russian}{спутник} & sputnik & fellow traveler/satellite/sputnik  \\\midrule
\foreignlanguage{russian}{ссылка} & ssylka & exile/link \\\midrule
\foreignlanguage{russian}{стол} & stol & table/diet  \\\midrule
\foreignlanguage{russian}{тачка} & tachka & wheelbarrow/car \\\midrule
\foreignlanguage{russian}{формат} & format & format\\\midrule
\foreignlanguage{russian}{центр} & tsentr & center \\\midrule
\foreignlanguage{russian}{ядро} & jadro & cannonball/core/nucleus \\
\bottomrule
    \end{tabular}
    \end{table*}

\end{document}